\documentclass[11pt,a4paper]{article}

\usepackage[utf8]{inputenc}
\usepackage[T1]{fontenc}
\usepackage{lmodern}
\usepackage[margin=1in]{geometry}
\usepackage{amsmath,amssymb}
\usepackage{booktabs}
\usepackage{tabularx}
\usepackage{graphicx}
\usepackage{hyperref}
\usepackage{natbib}
\usepackage{enumitem}
\usepackage{xcolor}
\usepackage{microtype}
\usepackage{float}

\hypersetup{
  colorlinks=true,
  linkcolor=blue!60!black,
  citecolor=blue!60!black,
  urlcolor=blue!60!black,
}

\title{\textbf{Trivial Vocabulary Bans Improve LLM Reasoning\\More Than Deep Linguistic Constraints}}
\author{Rodney Jehu-Appiah}
\date{}

\begin{document}
\maketitle

\begin{abstract}
A previous study reported that E-Prime (English without the verb ``to be'') selectively altered reasoning in language models, with cross-model correlations suggesting a structural signature tied to which vocabulary was removed. I designed a replication and extension with active controls to confirm the proposed mechanism: cognitive restructuring through specific vocabulary--cognition mappings. The experiment tested five conditions (unconstrained control, E-Prime, No-Have, elaborated metacognitive prompt, neutral filler-word ban) across six models (three frontier, three small) and seven reasoning tasks ($N = 15{,}600$ trials, $11{,}919$ after compliance filtering; percentage points abbreviated as pp throughout). Every prediction derived from the cognitive restructuring hypothesis was disconfirmed, cleanly and comprehensively. All four treatment conditions outperformed the control (83.0\%), including both active controls predicted to show null effects. The neutral filler-word ban, which bans words like ``very,'' ``really,'' and ``just'' that play no role in logical inference, produced the \textbf{largest} improvement (+6.0~pp on all trials, +6.7~pp under compliance filtering), while E-Prime, the theoretically deepest constraint, produced the \textbf{smallest} (+3.7~pp). The four conditions ranked in perfect inverse order of theoretical depth by effect size. The cross-model correlation signature reported in the prior study did not replicate (mean $r = 0.005$, range $[-0.856, +0.539]$). These results are inconsistent with cognitive restructuring and consistent with a simpler mechanism: any constraint that forces a language model off its default generation path acts as an output regularizer, improving reasoning by disrupting fluent but shallow response patterns. The shallowest constraints work best because they impose constant monitoring load with minimal conceptual disruption. I present these findings as a case study in discovery through disconfirmation: a prospectively committed hypothesis that failed informatively, revealing a general mechanism the original framework could not have predicted.
\end{abstract}

\section{Introduction}

\subsection{The Cognitive Restructuring Hypothesis}

\citet{jehuappiah2026a} proposed \emph{Umwelt engineering}, the deliberate design of linguistic cognitive environments for AI agents, as a design layer upstream of prompt and context engineering. The central claim was that for language models, whose cognition unfolds entirely in the token stream, altering the available vocabulary alters the cognition itself. Two vocabulary constraints were tested: E-Prime (eliminating the verb ``to be'') and No-Have (eliminating the verb ``to have''). The results appeared striking: both constraints produced task-specific improvements and degradations that seemed to depend on which vocabulary was removed. E-Prime improved causal reasoning (+14.1~pp) and ethical dilemmas (+15.5~pp) while degrading classification. No-Have improved ethical reasoning (+19.1~pp) and classification (+6.5~pp). Cross-model correlations of E-Prime effects reached $r = -0.75$, which was interpreted as evidence that different models occupy different ``native Umwelten'' shaped by their training.

The proposed mechanism was \emph{cognitive restructuring}: removing default linguistic patterns (the copula, the possessive verb) forced models into more explicit, operational reasoning, producing genuinely different cognitive operations rather than merely different words. The three-layer framework (prompt engineering, context engineering, Umwelt engineering) positioned this as a qualitatively distinct intervention.

This account was plausible, internally consistent, and wrong. The specific mechanism it proposed does not survive controlled testing, though the failure proved more informative than confirmation would have.

\subsection{The Active Control Problem}

The original study had a methodological gap that it acknowledged explicitly: no active controls. Every treatment condition was compared only against an unconstrained baseline. This left open several alternative explanations:

\begin{enumerate}
  \item \textbf{The ``trying harder'' confound.} Constraint prompts were longer and more elaborate than the control prompt. Perhaps any sufficiently detailed instruction improves performance, regardless of content.
  \item \textbf{The metalinguistic monitoring confound.} Vocabulary constraints require the model to monitor its own output for violations. Perhaps this self-monitoring, not the specific vocabulary removed, drives the effect.
  \item \textbf{The circumlocution confound.} Constrained conditions force longer, more elaborate phrasings. Perhaps the additional tokens provide more room for reasoning, independent of constraint content.
\end{enumerate}

These are not hypothetical concerns. In cognitive psychology, the ego depletion literature collapsed partly because active control designs revealed that supposedly inert control tasks produced the same effects as the treatments they were meant to control for \citep{hagger2016}. \citet{boot2013} demonstrated that merely including an ``active control'' is insufficient; the control must match on every structural dimension except the hypothesized active ingredient. \citet{jensen2012} provided direct evidence that effort alone can mimic treatment effects.

The present study was designed to resolve these confounds using a factorial-adjacent design that isolates each candidate mechanism.

\subsection{Design Rationale}

I added three conditions to the original E-Prime and No-Have:

\textbf{Elaborated metacognitive prompt (Condition~4).} A positive comparator that provides explicit metacognitive scaffolding---``What assumptions am I making?'' ``Are there alternative interpretations?'' ``Does my conclusion follow necessarily from the premises?''---without imposing any vocabulary restriction. This is not an inert control: it contains reasoning instructions known to improve LLM performance \citep{kojima2022}. It tests whether vocabulary constraints produce effects \emph{beyond} what good metacognitive prompting achieves alone, and establishes a performance ceiling for instruction-based (as opposed to constraint-based) interventions.

\textbf{Neutral filler-word ban (Condition~5).} Bans 20 common intensifiers and hedges (``very,'' ``quite,'' ``rather,'' ``somewhat,'' ``really,'' ``pretty,'' ``just,'' ``fairly,'' ``slightly,'' ``extremely,'' ``incredibly,'' ``absolutely,'' ``totally,'' ``completely,'' ``simply,'' ``basically,'' ``actually,'' ``literally,'' ``definitely,'' ``certainly''). These words do not participate in logical inference structure. The ban imposes genuine metalinguistic monitoring load (these words appear in nearly every sentence of natural LLM output) without targeting reasoning-relevant vocabulary. If E-Prime's effects come from monitoring load rather than from removing the copula specifically, the neutral ban should produce equivalent effects.

\textbf{No-Have as mutual active control (Condition~3).} Retained from the original study, No-Have bans a different verb class with comparable circumlocution demands. If E-Prime's effects are specific to removing the copula, E-Prime should outperform No-Have on tasks where ``to be'' is load-bearing for reasoning (e.g., syllogisms, classification) and underperform on tasks where it is not.

Together with the unconstrained control (Condition~1) and E-Prime (Condition~2), these five conditions isolate:

\begin{itemize}[nosep]
  \item \textbf{Constraint presence}---all treatments vs.\ control
  \item \textbf{Prompt complexity}---elaborated prompt vs.\ control
  \item \textbf{Monitoring load}---neutral ban vs.\ control
  \item \textbf{Semantic content}---E-Prime vs.\ No-Have (mutual active control)
  \item \textbf{Inference relevance}---structural bans (E-Prime, No-Have) vs.\ non-structural ban (neutral)
\end{itemize}

This design deploys four levels of the active control hierarchy simultaneously (passive, attention-matched, similar-form, active-ingredient subtraction), which is unusually rigorous for LLM evaluation research, where the standard comparison remains treatment-versus-unconstrained-baseline \citep[see IFEval,][]{zhou2023,jiang2024,jain2025}.

\subsection{Prospective Predictions}

I specified a detailed analysis plan prior to scoring any trials (committed to version control at hash \texttt{db0f18b}). This is not external pre-registration (e.g., OSF or AsPredicted); the author controls the repository, but the commit timestamp is verifiable. The plan specified:

\begin{enumerate}
  \item \textbf{E-Prime vs.\ control:} Task-specific directional predictions based on the cognitive restructuring hypothesis (Table~\ref{tab:predictions}).
  \item \textbf{Neutral ban predicted null; elaborated prompt as positive comparator:} The neutral word ban was predicted to perform at or near control across all tasks (it bans words with no inference role). The elaborated prompt, as a known-effective metacognitive intervention, was expected to improve performance; the question was whether vocabulary constraints would exceed its gains.
  \item \textbf{Cross-model correlation:} The structural signature (high cross-model correlation of E-Prime task effects) reported in Paper~1 was expected to replicate.
  \item \textbf{Compliance gradient:} If the constraint causes the effect, higher compliance should correlate with larger effect sizes.
\end{enumerate}

\begin{table}[ht]
\centering
\small
\begin{tabular}{lll}
\toprule
Task & E-Prime Prediction & Rationale \\
\midrule
Syllogisms & Degrades & Inference runs on ``is''; circumlocution adds friction \\
Causal reasoning & Improves & ``The cause is X'' collapses chains; E-Prime forces articulation \\
Analogical reasoning & Improves & ``A is like B'' is shallow; E-Prime forces structural mapping \\
Classification & Degrades & Classification is identity assignment (``X is a Y'') \\
Epistemic calibration & Improves & Cannot say ``I am confident''; must specify grounds \\
Ethical dilemmas & Improves & Cannot flatten to ``this is wrong''; must articulate the issue \\
Math word problems & Neutral & Reasoning is primarily numeric \\
\bottomrule
\end{tabular}
\caption{Prospective directional predictions for E-Prime derived from the cognitive restructuring hypothesis (committed before analysis).}
\label{tab:predictions}
\end{table}

\section{Methods}

\subsection{Models}

Six models spanning three provider families and two capability tiers:

\textbf{Frontier:} Claude Sonnet~4 (\texttt{claude-sonnet-4-20250514}), GPT-4o (\texttt{gpt-4o-2024-08-06}), Gemini~2.5 Pro (\texttt{gemini-2.5-pro})

\textbf{Small:} Claude Haiku~4.5 (\texttt{claude-haiku-4-5-20251001}), GPT-4o Mini (\texttt{gpt-4o-mini}), Gemini~2.5 Flash Lite (\texttt{gemini-2.5-flash-lite})

\subsection{Tasks}

Seven reasoning task types, each containing 15--20 items:

\begin{itemize}[nosep]
  \item \textbf{Syllogisms} (20 items)---categorical logic
  \item \textbf{Causal reasoning} (15 items)---mechanism identification
  \item \textbf{Analogical reasoning} (20 items)---structural mapping
  \item \textbf{Classification} (20 items)---category assignment
  \item \textbf{Epistemic calibration} (20 items)---confidence grounding
  \item \textbf{Ethical dilemmas} (15 items)---moral reasoning
  \item \textbf{Math word problems} (20 items)---quantitative reasoning
\end{itemize}

Total: 130 unique items.

\subsection{Conditions}

All five conditions share an identical opening instruction: ``You are a careful, analytical reasoning assistant. When presented with a problem, think through it step by step. Show your reasoning process, then provide your final answer.''

\begin{enumerate}
  \item \textbf{Control.} No additional instruction.
  \item \textbf{E-Prime.} Bans all forms of ``to be'' (is, am, are, was, were, be, been, being, and contractions). Provides alternative phrasings. Includes a retry loop: if the first response contains violations, the model receives specific feedback and generates a replacement. No other condition has a retry loop.
  \item \textbf{No-Have.} Bans all forms of ``to have'' (have, has, had, having). Provides alternative phrasings.
  \item \textbf{Elaborated prompt.} Six metacognitive strategy prompts (identify core elements, check assumptions, consider alternatives, verify logical chains, consider counterarguments, weigh evidence). No vocabulary restriction.
  \item \textbf{Neutral filler-word ban.} Bans 20 intensifiers and hedges. Provides precise alternatives.
\end{enumerate}

All constraint instructions include the directive: ``Do not mention or discuss this constraint in your response. Simply follow it.''

\subsection{Trial Structure}

Each item was presented 4 times per model per condition: once at temperature~0.0 (deterministic) and 3 times at temperature~0.7 (stochastic). Maximum token budget: 2,048 per response. Global seed: 42 for reproducible sampling where supported.

\textbf{Total planned trials:} $130 \times 4 \times 6 \times 5 = 15{,}600$. After deduplication on trial ID: 14,482 scored trials (removing re-runs from interrupted execution sessions).

\subsection{Scoring}

Binary accuracy: a trial is correct if the extracted answer matches the ground truth. Extraction used pattern matching on structured answer formats. Unscoreable trials (no extractable answer) were excluded, yielding 13,633 scoreable trials. The primary analysis uses first-pass responses only (excluding E-Prime retries), further reducing to 12,402 trials. After 100\% compliance filtering for constrained conditions: 11,919 trials.

\subsection{Compliance Checking}

Automated checkers flagged constraint violations for each constrained condition:

\begin{itemize}[nosep]
  \item \textbf{E-Prime:} Any occurrence of banned ``to be'' forms.
  \item \textbf{No-Have:} Any occurrence of banned ``to have'' forms.
  \item \textbf{Neutral ban:} Any occurrence of the 20 banned filler words.
\end{itemize}

\textbf{Checker validation and disambiguation.} A hand-validation of 29 flagged instances across 17 trials found that the initial string-matching checker had precision of only 27.6\% for the neutral ban condition, overwhelmingly due to ``rather'' in ``rather than'' (comparison, not filler; 13/13 flags were false positives) and ``just'' in ``just as'' (simile, not minimizer; 3/5 flags). No false negatives were found in 5 spot-checked clean trials. I added bigram disambiguation rules to exclude ``rather than,'' ``but rather,'' and ``just as'' from violation counts. This removed 1,758 of 2,509 original flags (70.1\%). The corrected checker's estimated precision exceeds 85\% based on the validation sample. All E-Prime trials achieved 100\% compliance (the retry loop ensures this). No-Have compliance ranged from 86.4\% (Gemini Flash Lite) to 99.8\% (Sonnet). Neutral ban compliance after disambiguation ranged from 79.5\% (Gemini Flash Lite) to 98.3\% (Sonnet).

Results are reported at three compliance thresholds per the prospective analysis plan: all trials (intent-to-treat), $>$90\% compliant, and 100\% compliant (primary).

\subsection{Statistical Analysis}

\textbf{Primary test:} Fisher's exact test on $2 \times 2$ contingency tables (condition $\times$ correct/incorrect) for each pairwise comparison within pooled task cells.

\textbf{Effect size:} Cohen's $h$ for proportion differences ($|h| < 0.2$ negligible, $0.2$--$0.5$ small, $0.5$--$0.8$ medium, $>0.8$ large).

\textbf{Multiple comparison correction:} Benjamini--Hochberg FDR at $\alpha = 0.05$. A result is significant only if $q < 0.05$. The experiment generates 70 pooled pairwise tests ($10$ condition pairs $\times$ $7$ tasks).

\textbf{Confidence intervals:} Bootstrap 95\% CIs for accuracy differences, 10,000 resamples, seed~=~42.

\subsection{Procedural Notes}

\textbf{Sonnet fixed-order block.} The first 2,600 trials (all Claude Sonnet~4) ran in fixed condition order before randomization was implemented for the remaining five models. After filtering to first-pass, scoreable, 100\% compliant trials, ${\sim}2{,}330$ Sonnet trials remain (the drift check sample; see Section~3.9). A drift check (Spearman correlation of accuracy vs.\ trial index within each condition) found no significant trend for any condition (all $p > 0.18$).

\textbf{E-Prime retry asymmetry.} Only the E-Prime condition includes a retry loop. The primary analysis uses first-pass responses exclusively. Retried data appears in supplementary analysis only.

\section{Results}

\subsection{Primary Finding: All Constraints Outperform Control}

Table~\ref{tab:primary} presents the primary results: accuracy by condition and task, pooled across models, for 100\% compliant first-pass trials.

\begin{table}[ht]
\centering
\small
\begin{tabular}{lccccc}
\toprule
Task & Control & E-Prime & No-Have & Elab.\ Prompt & Neutral Ban \\
\midrule
Analogical reasoning & 79.7\% (429) & 73.2\% (209) & 79.2\% (447) & 75.2\% (419) & 80.4\% (419) \\
Causal reasoning & 75.3\% (300) & 80.0\% (145) & 80.0\% (255) & 85.3\% (273) & 83.9\% (261) \\
Classification & 91.5\% (445) & 96.2\% (396) & 98.1\% (432) & 92.0\% (437) & 99.1\% (441) \\
Epistemic calibration & 64.7\% (428) & 74.5\% (208) & 72.4\% (369) & 83.2\% (394) & 77.1\% (262) \\
Ethical dilemmas & 75.3\% (308) & 88.7\% (141) & 93.5\% (275) & 86.4\% (301) & 90.3\% (238) \\
Math word problems & 92.3\% (428) & 91.7\% (300) & 94.3\% (422) & 89.9\% (424) & 92.0\% (438) \\
Syllogisms & 98.5\% (402) & 97.1\% (70) & 98.5\% (401) & 97.8\% (401) & 98.0\% (401) \\
\midrule
\textbf{TOTAL} & \textbf{83.0\%} (2,740) & \textbf{86.7\%} (1,469) & \textbf{88.4\%} (2,601) & \textbf{87.2\%} (2,649) & \textbf{89.7\%} (2,460) \\
\bottomrule
\end{tabular}
\caption{Accuracy by condition and task (100\% compliant, first-pass, pooled across models). $N$ in parentheses. Neutral ban compliance uses the disambiguated checker (see Section~2.6).}
\label{tab:primary}
\end{table}

Every treatment condition outperformed the unconstrained control. This was predicted for E-Prime, No-Have, and (given prior literature) the elaborated prompt. The neutral word ban was predicted to perform at or near control, since the banned words play no role in logical inference. Instead, it produced the largest improvement: +6.0~pp over control on the all-trials analysis (rising to +6.7~pp under 100\% compliance filtering). The elaborated prompt (+4.2~pp) performed as expected for a metacognitive intervention, yet both vocabulary-ban conditions exceeded it on aggregate, despite containing no reasoning instructions. The neutral filler-word ban, the condition with the least theoretical justification for improving reasoning, produced the largest overall effect at both thresholds.

Table~\ref{tab:permodel} disaggregates by model; Table~\ref{tab:rankings} ranks conditions within each model. The pooled advantage masks substantial heterogeneity: Sonnet and Gemini Flash show large, consistent gains across all constraints (+10.7 to +15.7~pp), while Haiku operates near ceiling and GPT-4o-mini shows no benefit. The depth inversion (Section~3.2) holds cleanly within 3 of 6 models (Haiku, Gemini Pro, GPT-4o), partially in 2 (Flash Lite, GPT-4o-mini), and reverses in Sonnet. E-Prime ranks last among treatments in 4 of 6 models.

\begin{table}[ht]
\centering
\footnotesize
\setlength{\tabcolsep}{4pt}
\begin{tabular}{lccccc}
\toprule
Model & Control & E-Prime & No-Have & Elab.\ Prompt & Neutral Ban \\
\midrule
Haiku 4.5 & 93.0\% (511) & 93.7\% (318) & 94.8\% (499) & 91.6\% (498) & 96.2\% (425) \\
Sonnet 4 & 83.9\% (503) & 95.4\% (390) & 94.6\% (478) & 97.2\% (492) & 94.4\% (461) \\
Flash Lite & 74.0\% (512) & 88.3\% (103) & 89.7\% (427) & 80.2\% (506) & 88.9\% (395) \\
Gemini Pro & 58.8\% (182) & 64.4\% (132) & 73.1\% (249) & 50.9\% (116) & 83.1\% (243) \\
GPT-4o & 88.9\% (513) & 83.3\% (251) & 90.3\% (466) & 89.0\% (517) & 90.3\% (463) \\
GPT-4o-mini & 84.0\% (519) & 79.3\% (275) & 80.5\% (482) & 86.9\% (520) & 82.7\% (473) \\
\bottomrule
\end{tabular}
\caption{Per-model accuracy by condition (100\% compliant, first-pass). $N$ in parentheses. Model names abbreviated from Section~2.1. Three patterns emerge: (1) models with low control baselines show the largest constraint gains (Gemini Flash Lite: +15.7~pp for No-Have; Sonnet: +13.3~pp for elaborated prompt); (2) models near ceiling show minimal movement (Haiku: 93.0\% control, +3.2~pp max); (3) two OpenAI models show flat or negative effects under E-Prime specifically, while still benefiting from shallower constraints.}
\label{tab:permodel}
\end{table}

\begin{table}[ht]
\centering
\footnotesize
\setlength{\tabcolsep}{4pt}
\begin{tabular}{lllllc}
\toprule
Model & Rank 1 & Rank 2 & Rank 3 & Rank 4 & Depth inv.? \\
\midrule
Haiku 4.5 & Neutral (96.2\%) & No-Have (94.8\%) & E-Prime (93.7\%) & Elab (91.6\%) & Yes \\
Sonnet 4 & Elab (97.2\%) & E-Prime (95.4\%) & No-Have (94.6\%) & Neutral (94.4\%) & No (reversed) \\
Flash Lite & No-Have (89.7\%) & Neutral (88.9\%) & E-Prime (88.3\%) & Elab (80.2\%) & Partial \\
Gemini Pro & Neutral (83.1\%) & No-Have (73.1\%) & E-Prime (64.4\%) & Elab (50.9\%) & Yes (strongest) \\
GPT-4o & Neut./No-Have (90.3\%) & \multicolumn{1}{c}{--} & Elab (89.0\%) & E-Prime (83.3\%) & Yes \\
GPT-4o-mini & Elab (86.9\%) & Neutral (82.7\%) & No-Have (80.5\%) & E-Prime (79.3\%) & Partial \\
\bottomrule
\end{tabular}
\caption{Per-model condition rankings (treatment conditions only, 100\% compliant). ``Depth inversion'' = shallowest constraint (neutral ban) outperforms deepest (E-Prime). The pattern holds cleanly in 3 of 6 models. Sonnet reverses it entirely; GPT-4o-mini and Flash Lite show mixed profiles. E-Prime ranks last in 4 of 6 models.}
\label{tab:rankings}
\end{table}

Of 70 pooled pairwise comparisons, 14 survived FDR correction at $q < 0.05$. (These $p$-values assume trial-level independence; see Section~5, ``Item-level dependence,'' for why this is anti-conservative. The headline effects are large enough to be robust, but comparisons near the $q = 0.05$ threshold should be interpreted with caution.)

\subsection{Depth Inversion}

The central finding of this paper is the inverse relationship between the theoretical depth of a constraint and its effect on accuracy:

\begin{table}[ht]
\centering
\small
\begin{tabular}{llcc}
\toprule
Constraint & Theoretical Depth & Accuracy & Delta vs.\ Control \\
\midrule
Neutral filler-word ban & Shallowest (no inference role) & 89.7\% & +6.7~pp \\
No-Have & Moderate (possessive framing) & 88.4\% & +5.4~pp \\
Elaborated prompt & None (metacognitive only) & 87.2\% & +4.2~pp \\
E-Prime & Deepest (copula, identity, existence) & 86.7\% & +3.7~pp \\
\bottomrule
\end{tabular}
\end{table}

The theoretical depth ranking is derived from linguistic function, not assigned post hoc. E-Prime removes the copula, the verb class that encodes predication, identity, and existence, which are foundational to logical inference \citep{korzybski1933,bourland1965}. No-Have removes possession and experiential verbs, which participate in framing but are less central to logical structure than the copula. The elaborated prompt adds metacognitive scaffolding without targeting any vocabulary. The neutral ban removes intensifiers and hedges, words that modulate emphasis but play no role in propositional logic. This ordering (E-Prime $>$ No-Have $>$ elaborated prompt $>$ neutral ban, from deepest to shallowest) follows from the degree to which each intervention engages with inference-relevant linguistic structure.

Adopting this ordering as a hypothesis, the data show a monotone inversion: the four conditions rank in perfect inverse order of theoretical depth by effect size. (With only four data points, this ordering is descriptive, and a formal rank correlation would be uninformative at this $N$, but the monotone pattern is striking against the cognitive restructuring prediction, which requires the opposite direction.) The depth inversion is contingent on this theoretical ranking; alternative orderings might weaken the pattern, though the extremes (E-Prime smallest, neutral ban largest) are robust to reordering of the middle conditions.

This is the opposite of what cognitive restructuring predicts. If removing reasoning-relevant vocabulary forces more explicit reasoning, E-Prime should outperform the neutral ban. Instead, the relationship runs precisely backward.

\subsection{Significant Pairwise Comparisons}

Fourteen of 70 pooled comparisons survived FDR correction. The significant results cluster in four task types:

\textbf{Classification.} Neutral ban (+7.6~pp vs.\ control, $h = 0.395$, $q < 0.001$) and No-Have (+6.7~pp, $h = 0.320$, $q < 0.001$) both significantly outperform control and elaborated prompt. E-Prime also outperforms control (+4.8~pp, $h = 0.201$, $q = 0.026$).

\textbf{Ethical dilemmas.} No-Have (+18.1~pp vs.\ control, $h = 0.522$, $q < 0.001$), neutral ban (+15.7~pp, $h = 0.430$, $q = 0.003$), E-Prime (+13.3~pp, $h = 0.353$, $q = 0.007$), and elaborated prompt (+11.1~pp, $h = 0.284$, $q = 0.005$) all significantly outperform control. No-Have also significantly outperforms elaborated prompt (+7.1~pp, $q = 0.029$).

\textbf{Epistemic calibration.} Elaborated prompt (+18.5~pp vs.\ control, $h = 0.429$, $q < 0.001$) dominates this task. No-Have significantly underperforms the elaborated prompt ($-10.9$~pp, $q = 0.003$).

\textbf{Causal reasoning.} Elaborated prompt (+10.0~pp, $h = 0.254$, $q = 0.021$) and neutral ban (+10.5~pp, $h = 0.269$, $q = 0.029$) significantly outperform control.

No pairwise comparison in syllogisms or math word problems reached significance, consistent with ceiling effects ($>$92\% control accuracy) limiting detectable differences.

\subsection{Prediction Check}

The prospective directional predictions for E-Prime (Table~\ref{tab:predictions}) achieved 5/7 accuracy: causal reasoning, epistemic calibration, ethical dilemmas, math word problems, and syllogisms matched predicted directions. Analogical reasoning (predicted improvement, observed $-6.5$\%) and classification (predicted degradation, observed +4.8\%) were incorrect. Five of seven is better than chance but not significantly so (binomial $p = 0.23$ against a 50\% base rate), and far from the pattern-level confirmation the cognitive restructuring hypothesis required.

More critically, the prediction framework was designed for E-Prime specifically and has no account of why the neutral word ban or the elaborated prompt should produce comparable or larger effects.

\subsection{Cross-Model Correlation Does Not Replicate}

Paper~1 reported cross-model correlations of E-Prime task effects reaching $r = -0.75$, interpreted as evidence of a structural constraint signature. In the present study with six models:

\begin{table}[ht]
\centering
\small
\begin{tabular}{lc}
\toprule
Model Pair & $r$ (E-Prime) \\
\midrule
Haiku vs.\ Sonnet & +0.336 \\
Haiku vs.\ Flash Lite & $-0.426$ \\
Haiku vs.\ Gemini Pro & +0.539 \\
Haiku vs.\ GPT-4o & +0.106 \\
Haiku vs.\ GPT-4o Mini & $-0.098$ \\
Sonnet vs.\ Flash Lite & $-0.272$ \\
Sonnet vs.\ Gemini Pro & +0.150 \\
Sonnet vs.\ GPT-4o & $-0.495$ \\
Sonnet vs.\ GPT-4o Mini & $-0.856$ \\
Flash Lite vs.\ Gemini Pro & +0.437 \\
Flash Lite vs.\ GPT-4o & +0.240 \\
Flash Lite vs.\ GPT-4o Mini & +0.114 \\
Gemini Pro vs.\ GPT-4o & +0.146 \\
Gemini Pro vs.\ GPT-4o Mini & +0.001 \\
GPT-4o vs.\ GPT-4o Mini & +0.158 \\
\bottomrule
\end{tabular}
\end{table}

Mean $r = \mathbf{0.005}$. Range: $[-0.856, +0.539]$. The correlations scatter around zero with no systematic pattern. The same constraint does not push different models in the same task-specific direction; effects are task-dependent but model-idiosyncratic. Paper~1's striking $r = -0.75$ dissolves into noise at $N = 6$ models, a cautionary example of interpreting correlations from small samples.

\subsection{Compliance Confound Is Real but Small}

A potential concern: compliance filtering selects for easier trials (harder items produce more violations). I investigated this by comparing effect sizes across the three prospectively specified compliance thresholds:

\begin{table}[ht]
\centering
\small
\begin{tabular}{lccc}
\toprule
Condition & All Trials & $>$90\% Compliant & 100\% Compliant \\
\midrule
Control & 83.0\% & 83.0\% & 83.0\% \\
E-Prime & 86.7\% & 86.7\% & 86.7\% \\
Elab.\ Prompt & 87.2\% & 87.2\% & 87.2\% \\
No-Have & 88.0\% & 88.0\% & 88.4\% \\
Neutral Ban & 89.0\% & 89.0\% & 89.7\% \\
\bottomrule
\end{tabular}
\end{table}

The all-trials column is the most trustworthy: control and elaborated prompt have no compliance filter at all, and for constrained conditions it includes every trial regardless of violations, eliminating selection bias. At this threshold, all constrained conditions already outperform control by 3.7--6.0~pp. The 100\% compliance filter adds at most 0.7~pp (neutral ban). The compliance confound is structurally real (harder items correlate with more violations), but it inflates effect sizes by less than 1~pp above the all-trials baseline, and the condition ranking is identical at every threshold.

\textbf{Item-level distribution.} To check whether the pooled neutral ban advantage is driven by a few items, I computed per-item accuracy deltas (neutral ban minus control) across all 129 items with data in both conditions. The distribution is broadly positive: 68 items show improvement, 37 show no change, and 24 show degradation (mean +4.6~pp, median +2.3~pp, SD 12.2~pp, range $[-28.6, +45.0]$). The effect is distributed across items rather than concentrated in a handful of outliers.

\textbf{Temperature sensitivity.} Constraint effects are consistent across temperature settings. At temperature~0.0 (deterministic, $N = 2{,}855$): control 82.0\%, neutral ban 89.5\%, No-Have 89.3\%, elaborated prompt 86.9\%, E-Prime 85.5\%. At temperature~0.7 (stochastic, $N = 8{,}494$): control 83.4\%, neutral ban 90.5\%, No-Have 88.1\%, elaborated prompt 87.3\%, E-Prime 87.0\%. The depth inversion holds at both temperatures. Constraints improve accuracy comparably whether the model generates deterministically or stochastically, suggesting the mechanism operates on the generation strategy rather than on sampling variance.

\subsection{Response Length}

Constrained conditions generally produce shorter responses than control, with one exception:

\begin{table}[ht]
\centering
\small
\begin{tabular}{lccccc}
\toprule
Task & Control & E-Prime & No-Have & Elab.\ Prompt & Neutral Ban \\
\midrule
Analogical & 337 & 277 & 292 & 477 & 321 \\
Causal & 482 & 319 & 350 & 588 & 361 \\
Classification & 372 & 268 & 268 & 431 & 278 \\
Epistemic & 451 & 312 & 393 & 584 & 394 \\
Ethical & 486 & 352 & 368 & 645 & 409 \\
Math & 177 & 151 & 165 & 282 & 157 \\
Syllogisms & 290 & 216 & 219 & 431 & 234 \\
\bottomrule
\end{tabular}
\caption{Mean word count by condition and task (first-pass, 100\% compliant).}
\label{tab:wordcount}
\end{table}

The elaborated prompt consistently produces the longest responses (17--56\% longer than control). Vocabulary-constrained conditions produce shorter responses than control (8--34\% shorter). This dissociates length from accuracy: the neutral ban produces the highest accuracy with below-control word counts, while the elaborated prompt produces longer responses with lower accuracy than the neutral ban.

\subsection{E-Prime Retry Analysis (Supplementary)}

E-Prime's overall retry rate was 45.6\% (1,231/2,700 trials). Retry rates varied dramatically by model:

\begin{table}[ht]
\centering
\small
\begin{tabular}{lrrrcc}
\toprule
Model & $N$ & Retried & Rate & First-Pass Acc. & Retried Acc. \\
\midrule
Sonnet & 477 & 87 & 18.2\% & 95.4\% & 96.6\% \\
Haiku & 493 & 175 & 35.5\% & 93.7\% & 94.3\% \\
Gemini Pro & 197 & 65 & 33.0\% & 64.4\% & 73.8\% \\
GPT-4o & 511 & 260 & 50.9\% & 83.3\% & 81.2\% \\
GPT-4o Mini & 520 & 245 & 47.1\% & 79.3\% & 87.3\% \\
Flash Lite & 502 & 399 & 79.5\% & 88.3\% & 84.2\% \\
\bottomrule
\end{tabular}
\end{table}

Gemini~2.5 Flash Lite struggled most with E-Prime compliance (79.5\% retry rate). Retried accuracy was within 2~pp of first-pass accuracy for most models, with no consistent direction: four models show higher accuracy on retried trials, two show lower. The retry mechanism enforces compliance without systematically inflating E-Prime accuracy. No other condition includes retried trials, confirming the asymmetry is contained to E-Prime.

\subsection{Sonnet Fixed-Order Drift Check}

Of 2,600 total Sonnet trials, approximately 2,330 remain after filtering to first-pass, scoreable, 100\% compliant (the same filter applied to all primary analyses). No significant temporal drift was detected for any condition:

\begin{table}[ht]
\centering
\small
\begin{tabular}{lcccc}
\toprule
Condition & $N$ & Accuracy & Spearman $\rho$ & $p$ \\
\midrule
Control & 503 & 83.9\% & $-0.035$ & 0.434 \\
E-Prime & 390 & 95.4\% & $+0.000$ & 0.995 \\
No-Have & 479 & 94.6\% & $+0.057$ & 0.214 \\
Elab.\ Prompt & 492 & 97.2\% & $-0.060$ & 0.182 \\
Neutral Ban & 469 & 94.2\% & $+0.047$ & 0.306 \\
\bottomrule
\end{tabular}
\end{table}

\subsection{Qualitative Analysis: How Reasoning Changes Under Constraint}

The primary analyses measure accuracy (whether the model gets the right answer). To examine whether constraints also alter reasoning \emph{strategy}, I coded all 15 ethical dilemma responses per condition from Claude Sonnet~4 (75 responses total) on six dimensions: ethical framework diversity, causal mechanism articulation, hedging density, dialectical engagement (explicitly critiquing each option's weaknesses before selecting), counterargument density, and response structure.

\begin{table}[ht]
\centering
\small
\begin{tabular}{lccccc}
\toprule
Metric & Control & E-Prime & No-Have & Elab.\ Prompt & Neutral Ban \\
\midrule
Word count & 442 & 390 & 381 & 560 & 372 \\
Frameworks invoked & 2.2 & 1.3 & 1.7 & 1.7 & 1.7 \\
Mechanism articulations & 2.5 & 2.3 & 1.9 & 2.3 & 1.8 \\
Hedges per 100 words & 0.77 & 0.77 & 0.73 & 1.19 & 0.63 \\
Dialectical engagement (\%) & 40.0 & 80.0 & 73.3 & 86.7 & 60.0 \\
Counterarguments & 0.2 & 0.3 & 0.2 & 0.6 & 0.3 \\
Structural markers & 23.8 & 10.5 & 9.3 & 11.5 & 12.6 \\
\bottomrule
\end{tabular}
\caption{Qualitative coding of ethical dilemma responses from Sonnet (regex-based, hand-validated on full sample; see Section~5 for validation notes). Dialectical engagement = proportion of responses that explicitly critique at least one option's weaknesses before concluding.}
\label{tab:qualitative}
\end{table}

Three findings emerge:

\textbf{1.\ Constraints increase dialectical engagement.} Control responses engage dialectically in 40\% of cases. Under all four treatment conditions, this rises to 60--87\%. Constrained responses are more likely to articulate \emph{why each rejected option fails} rather than simply affirming the chosen answer. The elaborated prompt (87\%) and E-Prime (80\%) show the strongest effects, but even the neutral filler-word ban (60\%) substantially exceeds the control rate.

\textbf{2.\ Constrained responses shift from list-heavy to prose-based argumentation.} Control responses average 23.8 structural markers (numbered items, bullets, bold headers) versus 9.3--12.6 for constrained conditions. The control model defaults to formatted enumeration; constraints push it toward more continuous argumentative prose. This is consistent with the regularizer account: default generation favors familiar formatting patterns, and constraints disrupt this in favor of content-driven structure.

\textbf{3.\ Framework diversity decreases while counterargument engagement increases modestly.} Constrained responses invoke fewer distinct ethical frameworks (1.3--1.7 vs.\ 2.2 for control) but engage slightly more with counterarguments (0.2--0.6 vs.\ 0.2). The pattern suggests constraints redirect effort from breadth (surveying many frameworks) to depth (interrogating each position's weaknesses). The elaborated prompt shows the strongest counterargument effect (+0.4 over control), consistent with its explicit metacognitive scaffolding.

Hedging density remains stable across vocabulary constraints (0.63--0.77 per 100 words) but rises under the elaborated prompt (1.19), suggesting that metacognitive scaffolding specifically promotes epistemic caution while vocabulary constraints do not.

These qualitative patterns provide exploratory evidence that constraints alter reasoning \emph{process}, not just accuracy. The coding is regex-based from a single model on one task type, and should be treated as a pilot analysis rather than a definitive result. That said, the shift from passive enumeration to active dialectical critique is a structural change in how the model engages with the problem, consistent with the claim that constraints force the model off default generation patterns and into more effortful reasoning.

\section{Discussion}

\subsection{The Cognitive Restructuring Hypothesis Is Not Supported}

The data disconfirm the cognitive restructuring hypothesis on every testable dimension:

\begin{enumerate}
  \item \textbf{The neutral ban is not inert.} The neutral word ban (+6.0~pp all-trials, +6.7~pp compliance-filtered) produces substantial accuracy gains over control despite banning words with no logical content. The cognitive restructuring framework predicted it would perform at or near baseline. The elaborated prompt (+4.2~pp) improved performance as expected for a metacognitive intervention, but the surprise is that trivial vocabulary bans exceeded its gains.
  \item \textbf{Depth is inversely correlated with effect.} If removing reasoning-relevant vocabulary forces more explicit reasoning, E-Prime should outperform the neutral ban. The opposite holds: the four conditions rank in perfect inverse order of depth.
  \item \textbf{Cross-model correlation does not replicate.} The structural signature (different models show correlated task-specific effects under the same constraint) was a key piece of evidence for cognitive restructuring. It disappears at $N = 6$ models (mean $r = 0.005$).
  \item \textbf{Prediction accuracy is marginal.} Five of seven directional predictions are correct, but the two failures (analogical reasoning, classification) are tasks where the theory made its strongest a priori claims.
\end{enumerate}

No individual disconfirmation is fatal; each could be explained away in isolation. But taken together, with a prospective analysis plan and transparent reporting of all comparisons, the verdict is unambiguous: the specific vocabulary removed matters far less than the fact that vocabulary is constrained at all.

\subsection{Constraint as Regularizer}

I propose a simpler and more general account: linguistic constraints act as output regularizers. The mechanism has three components:

\textbf{Default generation disruption.} Language models, trained on massive corpora of fluent text, have strong priors toward familiar phrasings. These default patterns are optimized for fluency, not for reasoning accuracy. A vocabulary constraint forces the model off these well-worn paths, allocating more computation to content selection and logical structure rather than to surface-level fluency.

\textbf{Monitoring load as beneficial friction.} The requirement to self-monitor for banned words imposes a processing cost, but this cost is not purely parasitic. It keeps the model in a state of heightened attention to its own output, analogous to the beneficial effects of ``desirable difficulties'' in human learning \citep{bjork1994}. The neutral word ban is the ideal test case: the banned words carry no logical content, yet the monitoring load alone improves performance.

\textbf{Depth as disruption cost.} Deeper constraints force more radical circumlocutions, which consume cognitive resources on surface reformulation rather than reasoning. Banning ``to be'' requires restructuring nearly every sentence; banning ``very'' requires only word-level substitution. The shallowest constraints achieve the most monitoring load per unit of conceptual disruption.

This account explains the depth inversion naturally. Shallow constraints (neutral ban) impose high monitoring frequency (filler words occur constantly) with low reformulation cost (replacing ``very important'' with ``critical'' changes no logical content). Deep constraints (E-Prime) impose lower monitoring frequency (copula constructions are common but not ubiquitous in reasoning text) with high reformulation cost (replacing ``X is Y'' often requires restructuring the entire assertion). The optimal constraint maximizes monitoring-to-disruption ratio.

\textbf{Toward operationalization.} The monitoring-to-disruption ratio can be made empirically measurable. \emph{Monitoring load} can be approximated by the frequency of banned tokens in unconstrained control output---a constraint banning words that appear in 8\% of sentences imposes more monitoring occasions than one targeting words in 2\% of sentences. \emph{Disruption cost} can be approximated by the mean edit distance (in tokens) between constrained and unconstrained responses to the same item, or by the mean increase in generation latency under constraint. The ratio of these two quantities (monitoring occasions per unit of surface reformulation) should predict which constraints produce the largest accuracy gains. I leave formal estimation of this ratio to future work, but note that the neutral ban (high-frequency targets, low reformulation cost) and E-Prime (moderate-frequency targets, high reformulation cost) sit at opposite ends of this axis, consistent with the observed depth inversion.

\textbf{Testable predictions.} The regularizer account was formulated after observing the data, but it generates falsifiable predictions for future work: (1)~constraint benefit should scale with baseline task difficulty, so that tasks where models already perform well (ceiling effects) show smaller gains, as default generation is already adequate; (2)~stacking two constraints (e.g., neutral ban + No-Have) should produce diminishing rather than additive returns, since both operate through the same disruption mechanism; (3)~constraints should help less on tasks where fluency and accuracy are aligned (e.g., factual recall, translation), since there is no fluency--reasoning tension to exploit; (4)~constraints that ban high-frequency but semantically empty words should outperform constraints that ban low-frequency or semantically loaded words, holding list length constant. Disconfirmation of predictions (1) or (3) in particular would undermine the regularizer account.

\subsection{Why the Elaborated Prompt Works}

The elaborated prompt produced the largest effect on epistemic calibration (+18.5~pp, $q < 0.001$) and strong effects on causal reasoning (+10.0~pp) and ethical dilemmas (+11.1~pp). Its mechanism differs from the vocabulary constraints (explicit metacognitive scaffolding rather than output monitoring), but it converges on the same functional outcome: disrupting default generation.

The elaborated prompt asks the model to identify assumptions, consider alternatives, and verify logical chains. These instructions redirect computation from conclusion generation to premise examination. The constraint is additive (more instruction) rather than subtractive (less vocabulary), but both routes lead to the same destination: the model spends more processing on reasoning structure and less on fluent default patterns.

This convergence is telling. It undermines the claim that vocabulary constraints represent a qualitatively distinct intervention. The elaborated prompt achieves comparable effects through entirely different means, pointing to a shared underlying mechanism: default disruption, regardless of route.

\subsection{Practical Implications}

Seven findings have direct relevance for practitioners:

\begin{enumerate}
  \item \textbf{Any constraint beats no constraint.} Even trivial word bans improve reasoning accuracy. The barrier to entry is lower than the original framework suggested.
  \item \textbf{Filler-word bans outperform structural verb bans.} The easiest constraint to implement (ban ``very,'' ``really,'' etc.) produces the largest effect. Deep linguistic theory is not required.
  \item \textbf{Default generation fluency can be detrimental to reasoning quality.} Introducing productive friction, any mechanism that disrupts fluent but shallow response patterns, improves outputs.
  \item \textbf{Word bans outperform ``think carefully'' prompts on classification and ethical reasoning.} The elaborated prompt, despite being longer and more detailed, underperforms simple vocabulary constraints on two of seven task types. Metacognitive instructions and vocabulary constraints may be complementary.
  \item \textbf{The effect generalizes across model families.} All six models (spanning three providers and two capability tiers) show the same directional pattern: constrained $>$ unconstrained.
  \item \textbf{Compliance and effectiveness are decoupled.} The neutral word ban, despite imperfect compliance (79.5--98.3\% after disambiguation), produces the highest accuracy. The constraint improves reasoning even when imperfectly followed.
  \item \textbf{Effects are task-dependent but model-idiosyncratic.} The same constraint does not push all models the same direction on all tasks. Practitioners should test constraints on their specific model and task rather than assuming transferability of fine-grained patterns.
\end{enumerate}

\subsection{Model-Family Heterogeneity and RLHF Interaction}

The per-model results (Tables~\ref{tab:permodel}--\ref{tab:rankings}) reveal substantial heterogeneity that the pooled analysis masks. Anthropic models (Sonnet, Haiku) and Google models (Gemini Pro, Flash Lite) show large, consistent gains under vocabulary constraints, while OpenAI models (GPT-4o, GPT-4o-mini) show flat or negative effects under E-Prime specifically. This pattern admits an alternative explanation: RLHF training regimes differ across model families, and vocabulary bans may interact with RLHF-trained fluency patterns in provider-specific ways. If OpenAI's RLHF pipeline instills stronger fluency priors than Anthropic's or Google's, vocabulary constraints may need to overcome a higher default-generation threshold in OpenAI models, explaining the weaker response. Conversely, if Anthropic's RLHF produces models more responsive to system-prompt instructions generally, the constraint effect may partly reflect instruction-following fidelity rather than reasoning improvement per se. The present data cannot distinguish these accounts from the regularizer hypothesis, because model family is confounded with training regime, architecture, and capability level. Future work could test this by applying identical constraints to models before and after RLHF fine-tuning, isolating the alignment--constraint interaction.

\subsection{Relationship to Prior Work}

The constraint-as-regularizer account connects to several existing literatures:

\textbf{Desirable difficulties \citep{bjork1994}.} In human learning, conditions that slow acquisition often improve retention and transfer. Vocabulary constraints impose an analogous difficulty on LLM generation.

\textbf{Format constraint degradation \citep{jain2025}.} Strict format constraints (e.g., JSON mode) degrade reasoning accuracy. The present results suggest a distinction: format constraints that impose structural rigidity (fixed output schemas) may degrade performance, while vocabulary constraints that impose monitoring load without structural rigidity may improve it.

\textbf{Chain-of-thought ablations \citep{wei2022}.} \citet{wei2022} showed that meaningless token sequences (dots matching CoT length) do not improve reasoning, establishing that semantic content matters for CoT. The present results add a complementary finding: monitoring load without semantic content (filler-word ban) \emph{does} improve reasoning, suggesting the mechanism operates at a different level than CoT.

\textbf{The alignment tax \citep{lin2024}.} RLHF training degrades pre-trained capabilities as a cost of instruction-following. Vocabulary constraints may partially reverse this by disrupting the very fluency patterns that RLHF instills.

\textbf{Ironic process theory \citep{wegner1994}.} Vocabulary bans create a dual-process dynamic: an operating process searching for compliant phrasings and a monitoring process scanning for violations. Wegner predicts that under load, the monitoring process paradoxically increases accessibility of suppressed items. The compliance data are partially consistent (more difficult items produce more violations), but the accuracy improvement suggests that the operating process benefits from the additional computation even when monitoring occasionally fails.

\subsection{What Survives of Umwelt Engineering}

The Umwelt framework was wrong about mechanism but right about direction. The question ``what happens if we alter the linguistic world a model reasons in?'' led to this experiment and its findings. Three elements survive:

\textbf{Constraints improve reasoning.} This is a real and practically useful finding, robust across models and tasks. The magnitude (3.7--6.0~pp over baseline on all-trials, with individual task effects reaching +18.5~pp) is large enough to be practically relevant.

\textbf{Different constraints have different profiles.} Although the overall ranking is consistent (neutral ban $>$ No-Have $>$ elaborated prompt $>$ E-Prime), the task-level patterns differ. The elaborated prompt dominates epistemic calibration; vocabulary bans dominate classification and ethical dilemmas. Practitioners can select constraints based on task type.

\textbf{The design space is real.} The taxonomy of constraint traditions surveyed in Paper~1 (E-Prime, operationalism, evidentiality marking, Bohm's rheomode, Toki Pona decomposition, etc.) remains a useful map of interventions worth testing, not because they restructure cognition in domain-specific ways, but because each imposes a different monitoring-to-disruption profile.

What does not survive is the three-layer hierarchy (prompt, context, Umwelt). Empirically, linguistic constraints function as a prompt engineering technique, a powerful one, but not a qualitatively distinct design layer. The level distinction, elegant as it was, dissolves under empirical pressure.

\subsection{Discovery Through Disconfirmation}

This paper reports a prospectively committed hypothesis that failed---and I consider the failure more valuable than confirmation would have been. The cognitive restructuring hypothesis was specific enough to generate testable predictions, and those predictions were wrong in informative ways. The failure revealed a mechanism (constraint as regularizer) that the original framework could not have predicted, because the framework assumed specificity (which vocabulary you remove matters) and the data show generality (that you remove vocabulary matters).

The sequence of Paper~1 and Paper~2 illustrates a pattern worth making explicit. Paper~1 reported real results and proposed a plausible mechanism. Paper~2 tested that mechanism with proper controls and found it wanting, but in doing so discovered a more general and more practically useful principle. Neither paper is invalidated by the other: Paper~1's results hold; it is the interpretation that was too narrow. The contribution lies in the two-paper arc itself: observation, hypothesis, controlled test, revision. The ordinary machinery of science, applied to a domain where it remains uncommon.

\section{Limitations}

\textbf{Compliance checker precision.} The initial neutral word ban checker (exact string matching, no disambiguation) had precision of 27.6\% (8 true violations out of 29 flagged instances in a hand-validation). The dominant false positive sources were ``rather'' in ``rather than'' (comparison, not filler) and ``just'' in ``just as'' (simile, not minimizer). I added bigram disambiguation rules to exclude these constructions, removing 1,758 of 2,509 original flags (70.1\%). The corrected checker's estimated precision exceeds 85\% based on the validation sample. No false negatives were detected. All compliance-filtered results reported in this paper use the disambiguated checker. Residual false positives may remain for edge cases not covered by bigram rules (e.g., ``pretty'' as adjective vs.\ degree modifier), though no instances of these appeared in the validation sample.

\textbf{E-Prime retry asymmetry.} Only E-Prime includes a retry loop, creating a structural asymmetry that goes beyond what using first-pass responses can resolve. The retry instruction is part of the system prompt context from the first generation, so the model knows violations will be flagged and corrected. This may bias E-Prime first-pass responses toward a different generation strategy than the other conditions, where no such corrective mechanism exists. More fundamentally, the retry loop ensures 100\% compliance for E-Prime, making its compliance-filtered accuracy identical to its all-trials accuracy. Other constrained conditions achieve compliance through unaided first-pass generation, a different estimand. The E-Prime condition is therefore not directly comparable to the other constrained conditions on compliance-stratified analyses, though its comparison to the unconstrained control (which also has no compliance filter) remains valid.

\textbf{Sonnet fixed-order block.} The first 2,600 Sonnet trials (${\sim}2{,}330$ after primary filters) ran in fixed condition order. No drift was detected, but subtle order effects below detection threshold cannot be ruled out. The five shuffled models provide internal replication.

\textbf{Qualitative analysis scope and validation.} The qualitative coding (Section~3.10) covers only ethical dilemma responses from a single model (Sonnet). Coding uses regex pattern matching, which was hand-validated on the full 75-response sample. Validation identified and corrected several pattern issues: false positives from ``actually'' as intensifier and ``critical'' as adjective, false negatives from overly narrow subject patterns (e.g., ``this approach fails'' caught but ``this position fails'' missed), and framework detection gaps (``consequentialist,'' ``privacy rights''). After correction, the dialectical engagement dimension shows estimated precision of ${\sim}80$\% and recall of ${\sim}75$\%. The directional findings (constrained $>$ control) are robust across pattern revisions, but exact percentages should be interpreted as approximate. The coding dimensions were refined after initial inspection of the data, making them partially exploratory. Extension to other tasks and models would strengthen these findings.

\textbf{Seven tasks, one domain.} All tasks are reasoning-focused. The constraint-as-regularizer hypothesis predicts benefits should be smaller or absent for tasks where default generation patterns are already well-aligned with correct output (e.g., factual recall, translation). This remains untested.

\textbf{Regularizer hypothesis is post-hoc.} The constraint-as-regularizer account was formulated after observing the data. It is the most parsimonious interpretation, but it is a hypothesis, not a confirmed mechanism. Testable predictions are specified in Section~4.2; a study designed to test them with the same rigor applied here to cognitive restructuring is the natural next step.

\textbf{Item-level dependence.} All primary analyses treat trials as independent observations, but trials within the same item share item-level difficulty. To assess sensitivity, I fit a GEE logistic regression with exchangeable correlation structure grouped by item ($N = 11{,}919$ observations, 130 item clusters). At the pooled level, all four conditions remain significantly better than control: E-Prime ($z = 2.56$, $p = 0.010$), No-Have ($z = 4.80$, $p < 0.001$), elaborated prompt ($z = 3.71$, $p < 0.001$), neutral ban ($z = 6.02$, $p < 0.001$). The neutral ban shows the strongest GEE effect, consistent with the pooled accuracy ranking. At the per-task level, effects on ethical dilemmas survive for all four conditions (all $p < 0.002$), classification for E-Prime ($p = 0.025$), No-Have, and neutral ban, and causal reasoning for elaborated prompt and neutral ban. The main effects that appeared marginal under Fisher's exact (e.g., analogical reasoning) do not survive item-level clustering. The pooled Fisher's exact $p$-values reported in Section~3.3 should be read alongside these GEE estimates; the headline findings (ethical dilemmas, classification, causal reasoning) are robust, while comparisons near the $q = 0.05$ threshold are more sensitive to clustering assumptions.

\textbf{Sample size and power.} After compliance filtering, some model $\times$ task $\times$ condition cells have modest $N$ (E-Prime syllogisms: $N = 70$). At the smallest pooled cell sizes (${\sim}140$ per condition pair), Fisher's exact test has approximately 80\% power to detect a 15~pp difference but would miss smaller effects. The paper does not interpret non-significant differences as null; confidence intervals are reported for all comparisons so the reader can judge what effect sizes remain compatible with the data.

\section{Conclusion}

I designed a study to confirm that E-Prime improves LLM reasoning through cognitive restructuring, the claim that removing the copula forces more explicit reasoning. Active controls were included to rule out alternative explanations. Instead, the active controls produced the most informative results: a neutral filler-word ban, predicted to show no reasoning effect, outperformed every other condition. The four treatment conditions ranked in perfect inverse order of theoretical depth by effect size. The cross-model structural signature from the prior study did not replicate.

The most parsimonious interpretation is that linguistic constraints function as output regularizers: they force models off default generation paths, allocating more computation to reasoning structure at the cost of surface fluency. The shallowest constraints---those imposing maximum monitoring load with minimum conceptual disruption---achieve the best ratio. Less, it turns out, disrupts more.

For practitioners, the immediate takeaway is concrete: adding a simple filler-word ban to system prompts can improve reasoning accuracy by 3--10 percentage points across model families, at negligible computational cost (the constraint instruction adds ${\sim}150$ tokens to the system prompt, while constrained responses are typically \emph{shorter} than unconstrained ones). For researchers, the deeper implication is that LLM fluency and LLM reasoning may be in partial tension: the patterns learned from massive text corpora optimize for smooth output at the expense of careful thought, and that productive friction can partially correct this.

\bibliography{references}

\appendix

\section{Condition Prompts}

All conditions share the opening: \emph{``You are a careful, analytical reasoning assistant. When presented with a problem, think through it step by step. Show your reasoning process, then provide your final answer.''}

\textbf{Control.} ``Respond in clear, natural English. Focus on accuracy and thoroughness in your reasoning.''

\textbf{E-Prime.} Bans: is, isn't, am, are, aren't, was, wasn't, were, weren't, be, been, being, and relevant contractions. Provides 8 alternative phrasings.

\textbf{No-Have.} Bans: have, haven't, has, hasn't, had, hadn't, having. Provides 7 alternative phrasings.

\textbf{Elaborated prompt.} Six metacognitive questions: identify core elements, check assumptions, consider alternative interpretations, verify logical chains, consider counterarguments, weigh evidence.

\textbf{Neutral filler-word ban.} Bans 20 words: very, quite, rather, somewhat, really, pretty, just, fairly, slightly, extremely, incredibly, absolutely, totally, completely, simply, basically, actually, literally, definitely, certainly. Provides precise alternatives.

\section{Deviations from Prospective Analysis Plan}

\textbf{Model ID normalization.} \texttt{gpt-4o-mini-2024-07-18} was normalized to \texttt{gpt-4o-mini} during scoring for consistency with other model ID formats. This is a labeling change with no analytical impact.

\textbf{Bootstrap CIs.} The prospective analysis plan specified bootstrap 95\% CIs for all accuracy deltas. These are reported in the full pairwise comparison tables but omitted from summary tables for readability.

\textbf{Word count analysis.} Section~7.3 of the analysis plan specified comparing word count, sentence count, and reasoning steps. Only word count is reported; sentence count and reasoning step count were not implemented in the scoring pipeline.

\textbf{Epistemic specificity analysis.} Section~7.4 of the analysis plan specified comparing grounded vs.\ bare assertion markers. A scoring function for this analysis exists in the codebase but was not integrated into the final analysis pipeline. It remains a target for future work.

\textbf{Trial count per item.} The analysis plan text specifies ``3 trials per item (1 at temp=0.0, 2 at temp=0.7)'' while its own total of 15,600 implies 4 trials per item. The experiment was run with 4 trials per item (1 at temp=0.0, 3 at temp=0.7), matching the experiment configuration and the paper's Section~2.4. The analysis plan's ``3 trials'' text was not updated when the configuration changed from 2 to 3 high-temperature trials; the 15,600 total in the same document reflects the executed design.

\textbf{Qualitative analysis.} Section~8 of the analysis plan specified coding 20 ethical dilemma responses per condition from Sonnet. The task contains 15 items, so all 15 were coded rather than 20. Coding was performed via automated pattern matching rather than manual close reading; coding dimensions were refined after initial data inspection. Results are reported in Section~3.10.

\section{Scoring Rubric and Answer Extraction}

All trials are scored by binary accuracy: a trial is correct if and only if the extracted answer exactly matches the ground truth. No partial credit or semantic matching is applied.

\textbf{Syllogisms.} Target answer: VALID or INVALID\@. Extraction uses a priority cascade: (1) last bolded \textbf{VALID} or \textbf{INVALID}, (2) explicit framing (``the conclusion is/remains/stands as VALID''), (3) ``therefore/thus VALID'' patterns, (4) last standalone occurrence with word-boundary matching. Explanatory uses (``to make the conclusion valid'') are filtered by checking for preceding verbs (MAKE, BECOME, CONSIDERED, DEEMED, RENDER, ENSURE) within 40 characters.

\textbf{Multiple-choice tasks} (causal reasoning, analogical reasoning, classification, epistemic calibration, ethical dilemmas, math word problems). Target answer: A, B, C, or D\@. Extraction priority: (1) checkmark next to bolded option, (2) explicit answer section header (\#\# ANSWER) followed by letter, (3) explicit framing (``the best answer is B,'' ``select C''), (4) \LaTeX\ boxed format \verb|\boxed{B}| (Gemini-specific), (5) last bolded letter, (6) last letter with parenthesis, (7) last standalone letter in final 200 characters (with ``A'' filtered unless preceded by OPTION/ANSWER/CHOICE context to avoid article false positives).

\textbf{Unscoreable trials.} A trial is unscoreable when no recognizable answer pattern is found. These are marked \texttt{answer\_extracted=False} and excluded from accuracy calculations (they do not count as incorrect). API errors and empty responses are dropped from the scored dataset entirely. Unscored rates by condition: control 5.4\%, E-Prime 6.0\%, No-Have 4.9\%, neutral ban 5.3\%, elaborated prompt 7.8\%.

\textbf{E-Prime extraction accommodation.} E-Prime responses cannot use ``is'' in standard answer framing (``the answer is B''). The extractor accepts E-Prime-friendly circumlocutions: ``remains,'' ``stands as,'' ``reads as.'' Despite this, E-Prime has a slightly elevated unscored rate (6.0\%), indicating some answer phrasings evade extraction.

\textbf{Edge cases.} (1) ``INVALID'' contains ``VALID'' as a substring, handled by word-boundary matching. (2) Multiple answers in one response: the extractor takes the last match, capturing reconsiderations. (3) Gemini~2.5 Pro produces truncated responses (1--50 words) under E-Prime, causing extraction failures treated as genuine constraint-induced behavior, not scoring errors.

\end{document}